\documentclass[journal]{IEEEtran}
\usepackage[export]{adjustbox}
\usepackage{color, colortbl}
\usepackage{xcolor}

\usepackage[]{todonotes}

\newcommand{\statement}[1]{\todo[inline,color=gray!30]{#1}}

\usepackage{mdframed}
\newmdenv[innerlinewidth=0.5pt,innerleftmargin=6pt,
innerrightmargin=6pt,innertopmargin=6pt,innerbottommargin=6pt,roundcorner=4pt,backgroundcolor=gray!30]{roundbox}

\newcommand{\subparagraph}{}

\usepackage{cite}

\usepackage{amsmath}
\usepackage{gensymb}

\ifCLASSOPTIONcompsoc
  \usepackage[caption=false,font=normalsize,labelfont=sf,textfont=sf]{subfig}
\else
  \usepackage[caption=false,font=footnotesize]{subfig}
\fi

\usepackage{stfloats}

\usepackage{url}
\begin{document}
%
\title{Grasp success prediction with quality metrics}

\author{Carlos Rubert, Daniel Kappler, Jeannette Bohg and Antonio Morales
\thanks{C. Rubert and A. Morales are with the Robotic Intelligence Laboratory at the Department
of Computer Science and Engineering, Universitat Jaume I of Castellon, Spain, e-mail: cescuder,morales@uji.es.}
\thanks{D. Kappler is with the Autonomous Motion Department at the Max-Planck Institute for Intelligent Systems, Germany, e-mail: 
daniel.kappler@tuebingen.mpg.de.}
\thanks{J. Bohg is with the Department of Computer Science, Stanford University, USA, e-mail: bohg@cs.stanford.edu.}}


\maketitle


\begin{abstract}
Current robotic manipulation requires reliable methods to predict
whether a certain grasp on an object will be successful or
not prior to its execution. Different methods and metrics have been 
developed for this purpose but there is still work to do to provide a robust
solution.

In this article we combine different metrics to
evaluate real grasp executions. We use different machine learning algorithms  to train a classifier
able to predict the success of candidate grasps. 
Our experiments are performed with two different robotic grippers and
different objects. Grasp candidates are evaluated in both
simulation and real world.

We consider 3 different categories to label grasp executions: robust, fragile and futile. Our results shows the proposed prediction model has success rate of 76\%.

\end{abstract}



\section{Introduction}

In this article we analyze how well different, commonly-used grasp metrics are able to predict grasp success, either individually or in combination. In a previous study \cite{Rubert2017b}, it was seen that these metrics capture different aspects of precision grasp stability and thereby have different biases and generalization characteristics. Results in this study showed machine learning algorithms were able to find a non-trivial mapping from each of these metrics or a combination of them to the binary decision of grasp success. 

Our aim is to find a classifier that takes one or all of these metrics as input and outputs a success label. We use real experiments to test a set of classification methods which would lead to a prediction model for simulated grasps.

\begin{figure}[htbp]
\subfloat{\includegraphics[width=0.25\textwidth]{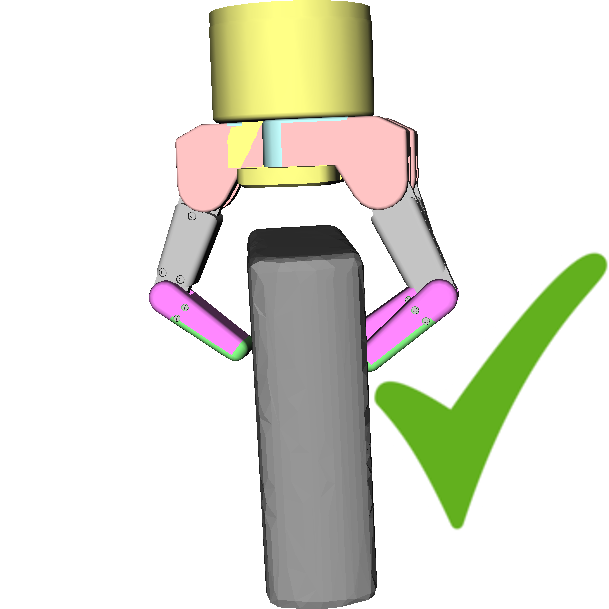}}
\subfloat{\includegraphics[width=0.20\textwidth]{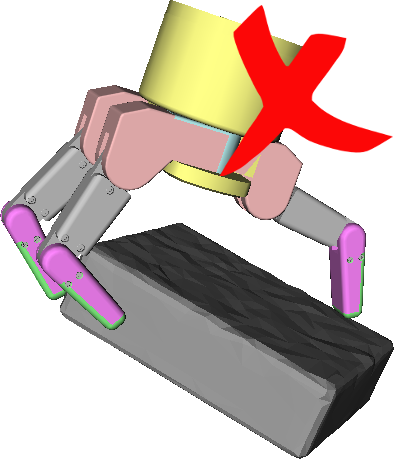}}
\\
\subfloat{\includegraphics[width=0.25\textwidth]{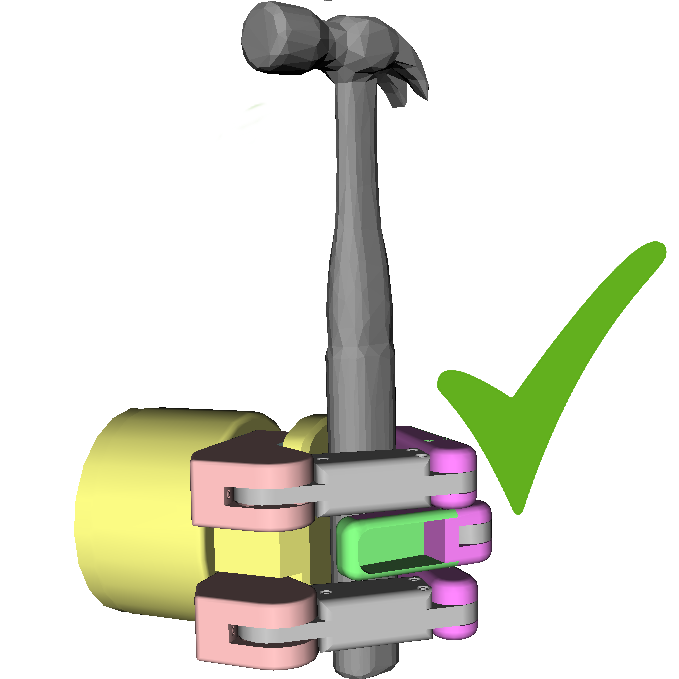}}
\subfloat{\includegraphics[width=0.25\textwidth]{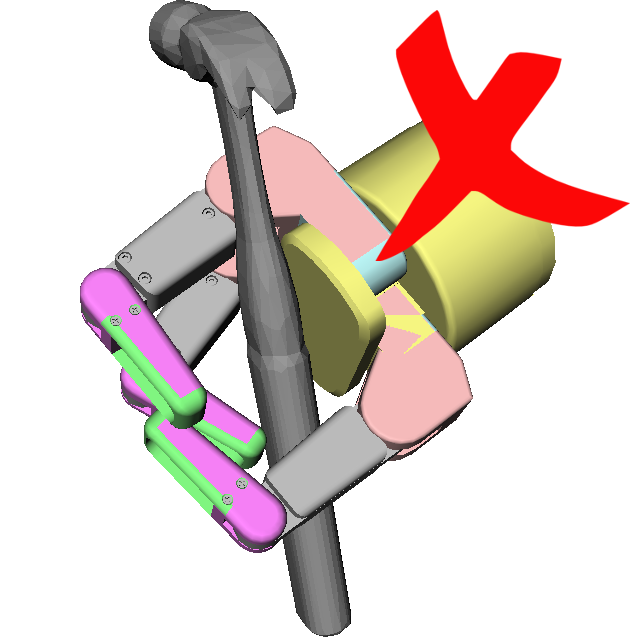}}
\\
\caption{Examples of \textit{Successful} (green mark) and {\em Unsuccessful}
  (red mark) grasps.}
\label{fig:grasps_success}
\end{figure} 

We employ a 3-grade scale to label grasp executions as \textit{Robust}, \textit{Futile} or \textit{Fragile}. A grasp is considered to be \textit{robust} when its executions always succeed. A grasp is considered \textit{futile} when its executions always fail. A grasp is fragile when its executions could fail or succeed, meaning its highly dependent of achieving accurate contact points.

\section{Quality metrics}
\label{sec2}

This study relies in the use of quality metrics to evaluate grasp executions. In a previous study of the authors, \cite{Rubert2018}, is performed a statistical analysis of the values produced by ten selected quality metrics on a database that included grasps for seven different hands and more than hundred
objects, resulting in around 900.000 different grasp configurations. The analysis is performed exclusively in simulation and consisted in establishing upper and lower thresholds for the
normalization of the metrics; measuring the stability of the metrics in the presence of small disturbances; and more importantly, visualizing the correlation between different metrics. This last result allowed to discard three metrics reducing the initial set to seven. These are the seven independent metrics which will be used in this article (see Table \ref{tab:ind_qm}). 

\begin{table}[h]
 \caption{Summary of the selected independent quality metrics}
 \label{tab:ind_qm}
 \resizebox{\columnwidth}{!}{%
 \renewcommand{\arraystretch}{1.5}
 \setlength\tabcolsep{5pt}
 \begin{tabular}{l p{4.5cm} >{\centering\arraybackslash}m{3cm}}
 	\hline
 	Name &  & Formula \\
 	\hline
 	$Q_{A1}$ & Smallest singular value of $G$ \cite{Li1987}  & \scalebox{1.1}{$\sigma_{min}(G)$} \\
 	\noalign{\smallskip}
 	$Q_{B1}$ & Distance between the centroid of the contact polygon and the center
 of mass of the object \cite{Ding2001,Ponce1997}  & \scalebox{1}{$1-\dfrac{\textit{distance}(p, p_c)}{\textit{distance}_{max}}$} \\
 	$Q_{B2}$ & Area of the grasp polygon \cite{Mirtich1994}  & \scalebox{1}{$\dfrac{\textit{Area}(\textit{Polygon}(p_1,...p_n))}{Area_{max}}$} \\
 	$Q_{B3}$ & Shape of the grasp polygon \cite{Kim2001}  &
 	\scalebox{1.1}{$1-\dfrac{1}{\theta_{max}} \sum\limits_{i=1}^{n_f} |\theta_i - \bar{\theta}|$} \\
 	\noalign{\smallskip}
 	$Q_{C2}$ & Volume of the convex hull \cite{Miller1999}  & \scalebox{1}{$\dfrac{\textit{Volume}(CW)}{\textit{Volume}_{max}}$} \\
 	\noalign{\smallskip}
 	$Q_{D1}$ & Posture of manipulator joints \cite{Liegeois1977}  & \scalebox{1.1}{$1 - \dfrac{1}{n_q} \sum\limits_{i=1}^{n_q} (\dfrac{y_i-a_i}{a_i-y_{iM}})^2$} \\
 	\noalign{\smallskip}
 	$Q_{D2}$ & Inverse of the condition number of $\mathbf{G_J}$ \cite{Salisbury1982a,Kim1991}  & $\dfrac{\sigma_{min}(G_J)}{\sigma_{max}(G_J)}$ \\
 	\noalign{\smallskip}
 	\hline
 	\end{tabular}
 	}
 \end{table}

The results of these works offer numerical and practical information
about the use of the metrics. The main motivation of
 this article is to address the limitation of this study that no relation between highly ranked grasps and the success of their execution is presented.

\section{Classification methods}
\label{sec2:clas}

We aim to find a model $y = f(\mathbf{x}; \mathbf{w})$ that can predict binary grasp success $y$ given an input feature vector $\mathbf{x}$ consisting of different grasp quality metrics. Our approach is to learn a classifier from the data obtained through real experiments that ideally minimizes the following equation:  

\begin{align}
  \underset{\mathbf{w}} {\min \,}
  \sum_{(\mathbf{x},y) \in \mathcal{D}} 1- l(f(\mathbf{x}; \mathbf{w}), y)
  \label{eq:loss_general}
\end{align}

where $\mathbf{w}$ denotes the parameter vector of the classifier
and
\begin{align}
  l(f(\mathbf{x}; \mathbf{w}), y) =
  \begin{cases}
    1, & \text{if } f(\mathbf{x}; \mathbf{w}) = y\\
    0, & \text{otherwise}
  \end{cases}
\end{align}

Every classifier considered in this article minimizes a loss in this setting. For various reasons, the exact loss formulations may vary per method, e.g. through different regularizers or by dropping the indicator function in order to get gradients. 

Given this data set $\mathcal{D}$, we train two different classifiers
using SciKit-Learn \cite{scikit-learn}: \textit{classification trees} and \textit{k-nearest neighbours}. The optimization for each method was done using grid search and cross-validation. Details are presented in each subsection.  A more in-depth description of the classification methods used in this chapter can be found in~\cite{Bishop2006}. 

\section{Materials}

\subsection{Robotic platforms}
\begin{figure}[htb]
\center
\captionsetup[subfloat]{width=3.5cm}
\subfloat[Barrett hand]{\includegraphics[width=0.5\columnwidth]{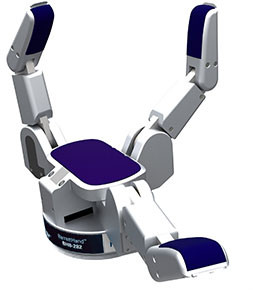}}
\captionsetup[subfloat]{width=3.5cm}
\subfloat[Schunk SDH hand]{\includegraphics[width=0.45\columnwidth]{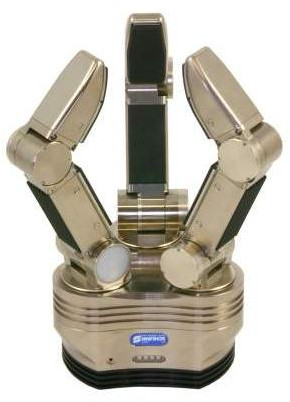}}
\caption{Real models of the robotic grippers.}
\label{fig:real_grippers}
\end{figure}

Our experiments consider two different robotic platforms: Apollo\footnote{Apollo Robot: \url{https://am.is.tuebingen.mpg.de/pages/robots}} and Tombatossals\footnote{Tombatossals Robot: \url{http://robinlab.uji.es/our_robots}} or \textit{tombato}. The \textit{Barrett} (Tombatossals) and \textit{Schunk SDH} (Apollo) grippers (Figure \ref{fig:real_grippers}) are used for performing the experiments with real robots. In both models the closure of the fingers is done until a contact is detected. Then, the joint is blocked and the closure continues for the distal joint. For the Barrett hand, a strain measurement detects the collision in the finger. In the case of the Schunk SDH, there are tactile sensors for detecting the contact on each link of the hand.

\subsection{Objects}

\begin{figure}[h]
\center
\captionsetup[subfloat]{width=3cm}
\subfloat[Bottle\_050]{\includegraphics[width=.32\columnwidth]{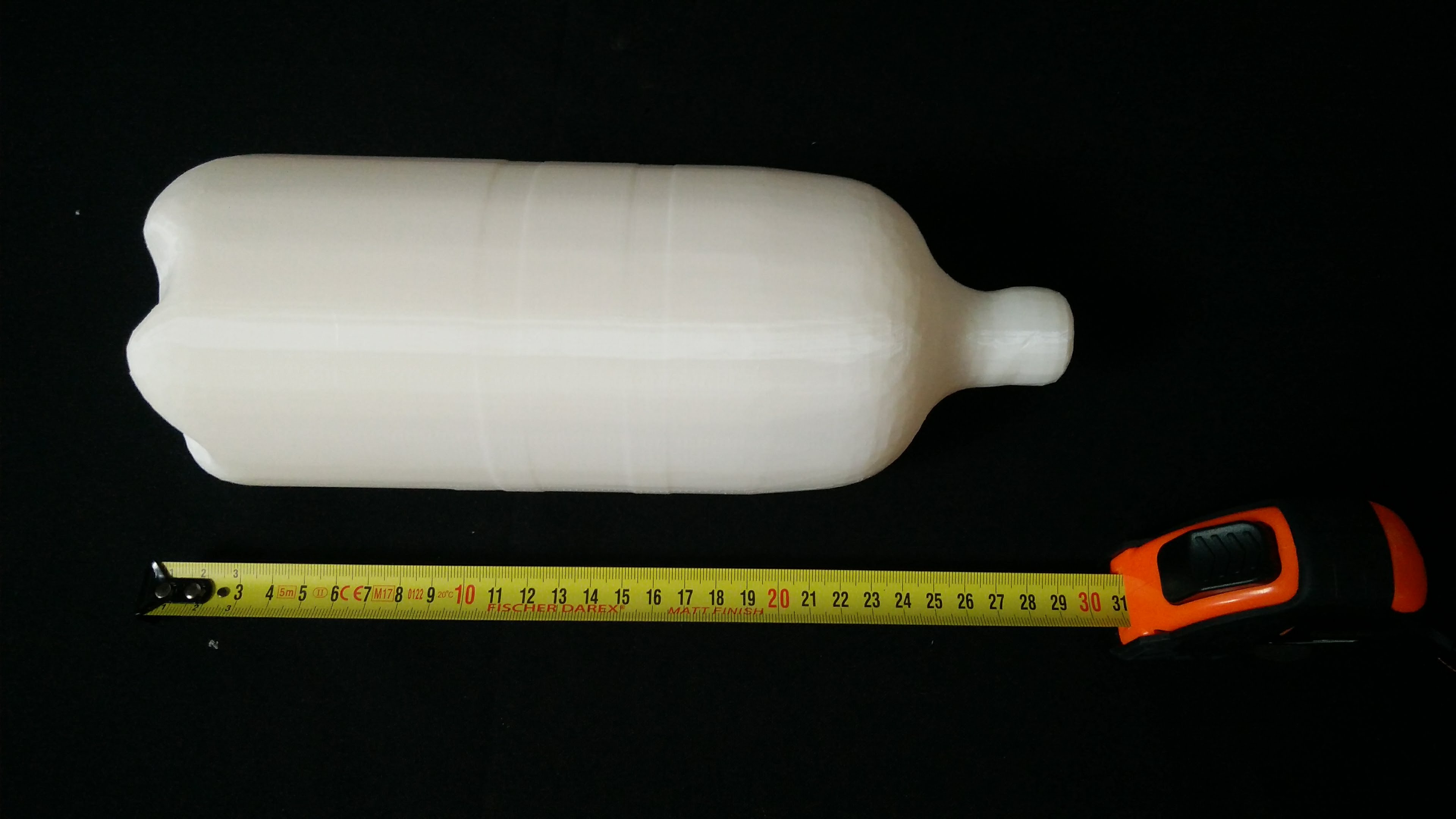}}
\captionsetup[subfloat]{width=3cm}
\subfloat[Bottle\_047]{\includegraphics[width=.32\columnwidth]{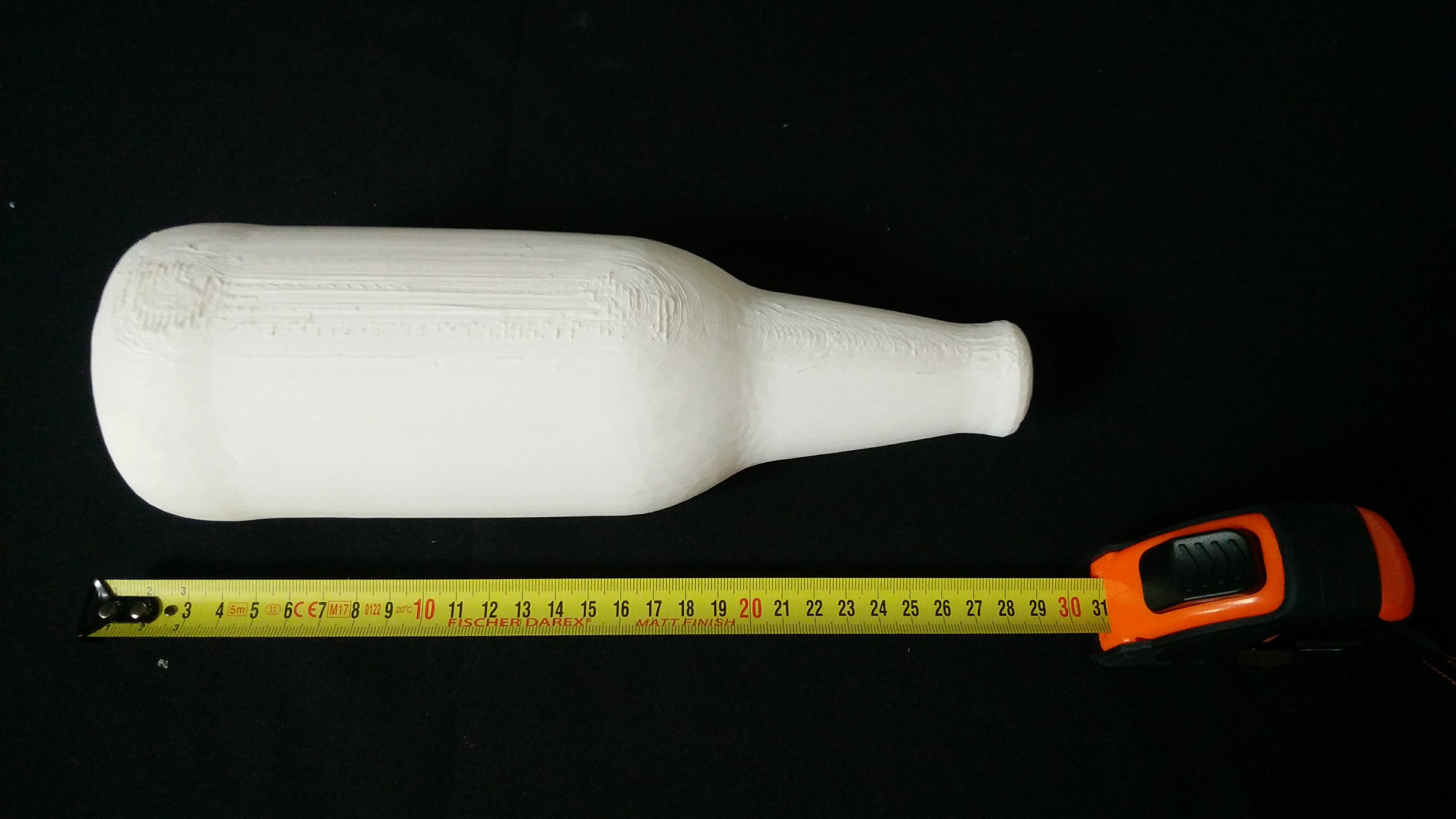}}
\captionsetup[subfloat]{width=3cm}
\subfloat[Camera\_015]{\includegraphics[width=.32\columnwidth]{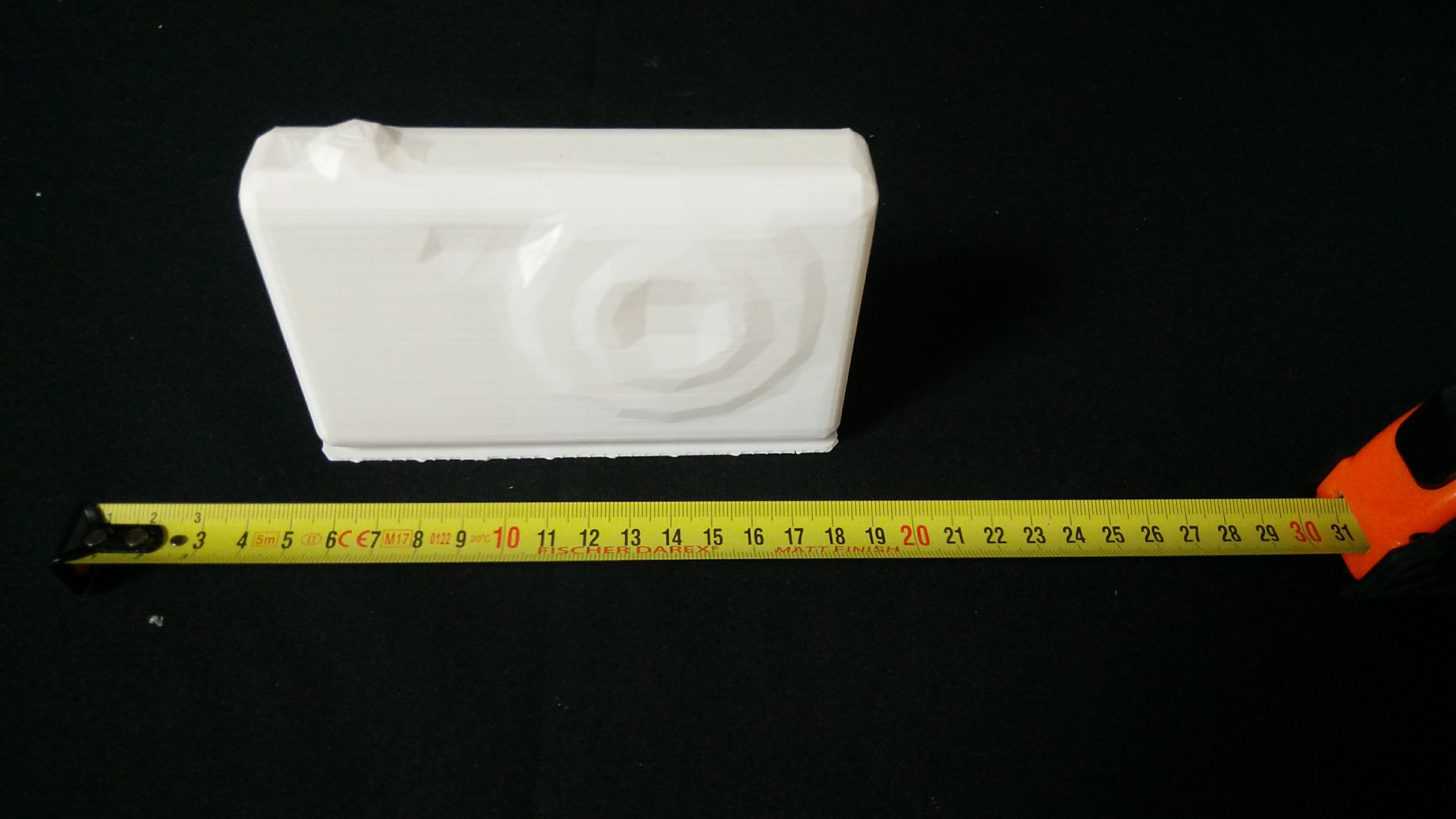}}
\\
\captionsetup[subfloat]{width=3cm}
\subfloat[][Lemon\_003]{\includegraphics[width=.32\columnwidth]{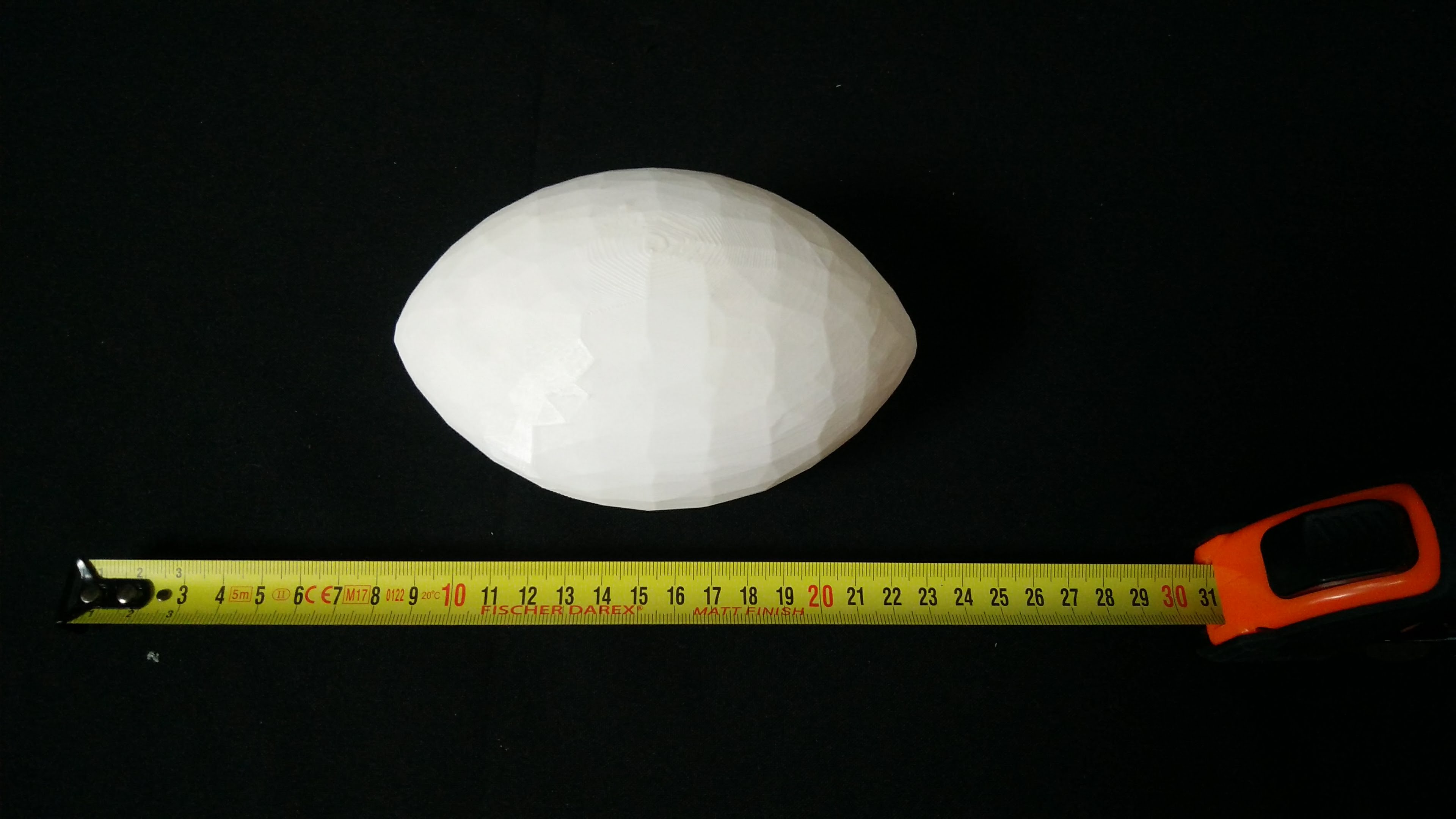}}
\captionsetup[subfloat]{width=3cm}
\subfloat[][Bowl\_022]{\includegraphics[width=.32\columnwidth]{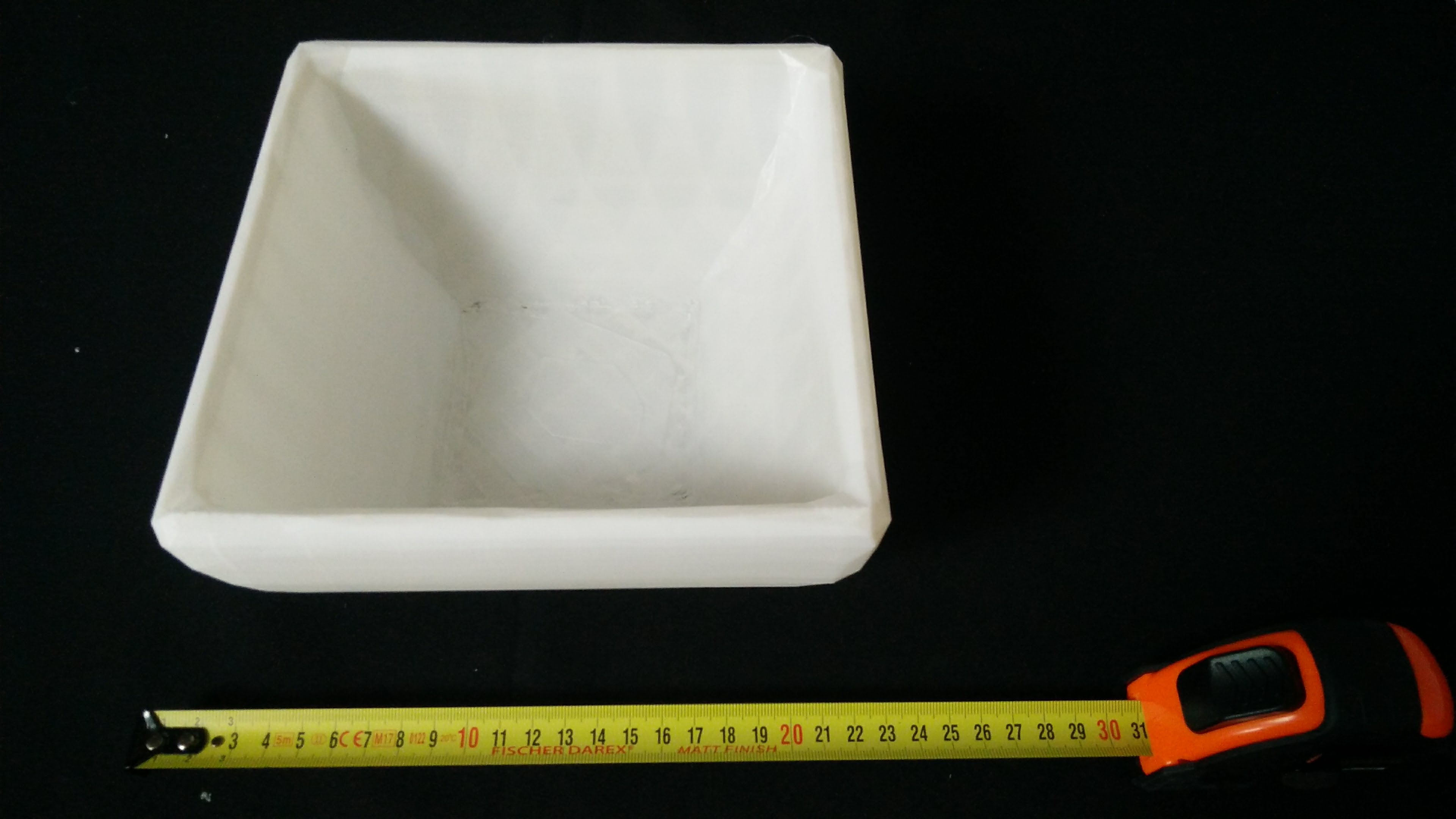}}
\captionsetup[subfloat]{width=3cm}
\subfloat[][Bowl\_025]{\includegraphics[width=.32\columnwidth]{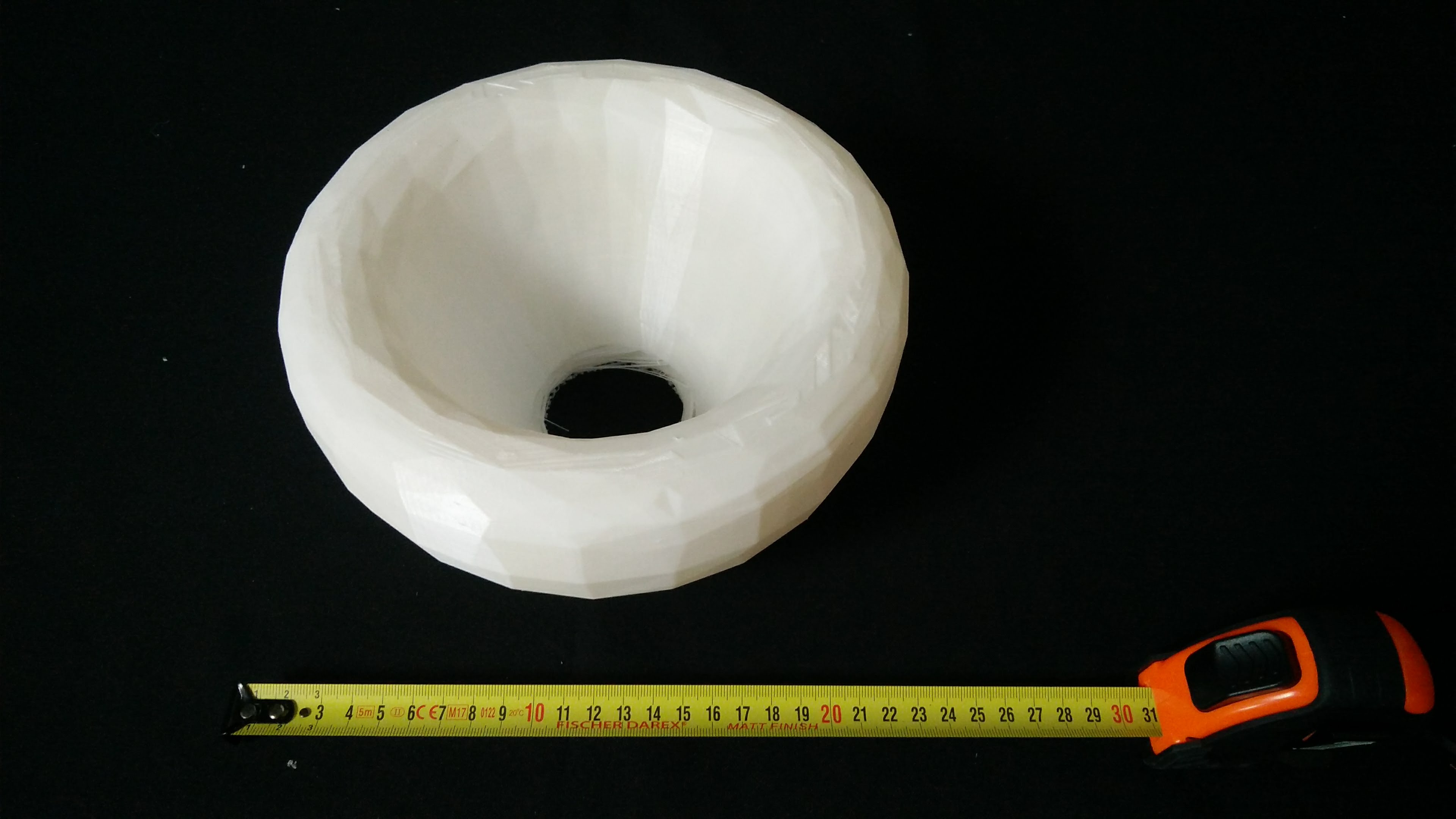}}
\\
\captionsetup[subfloat]{width=3cm}
\subfloat[][Toaster\_001]{\includegraphics[width=.32\columnwidth]{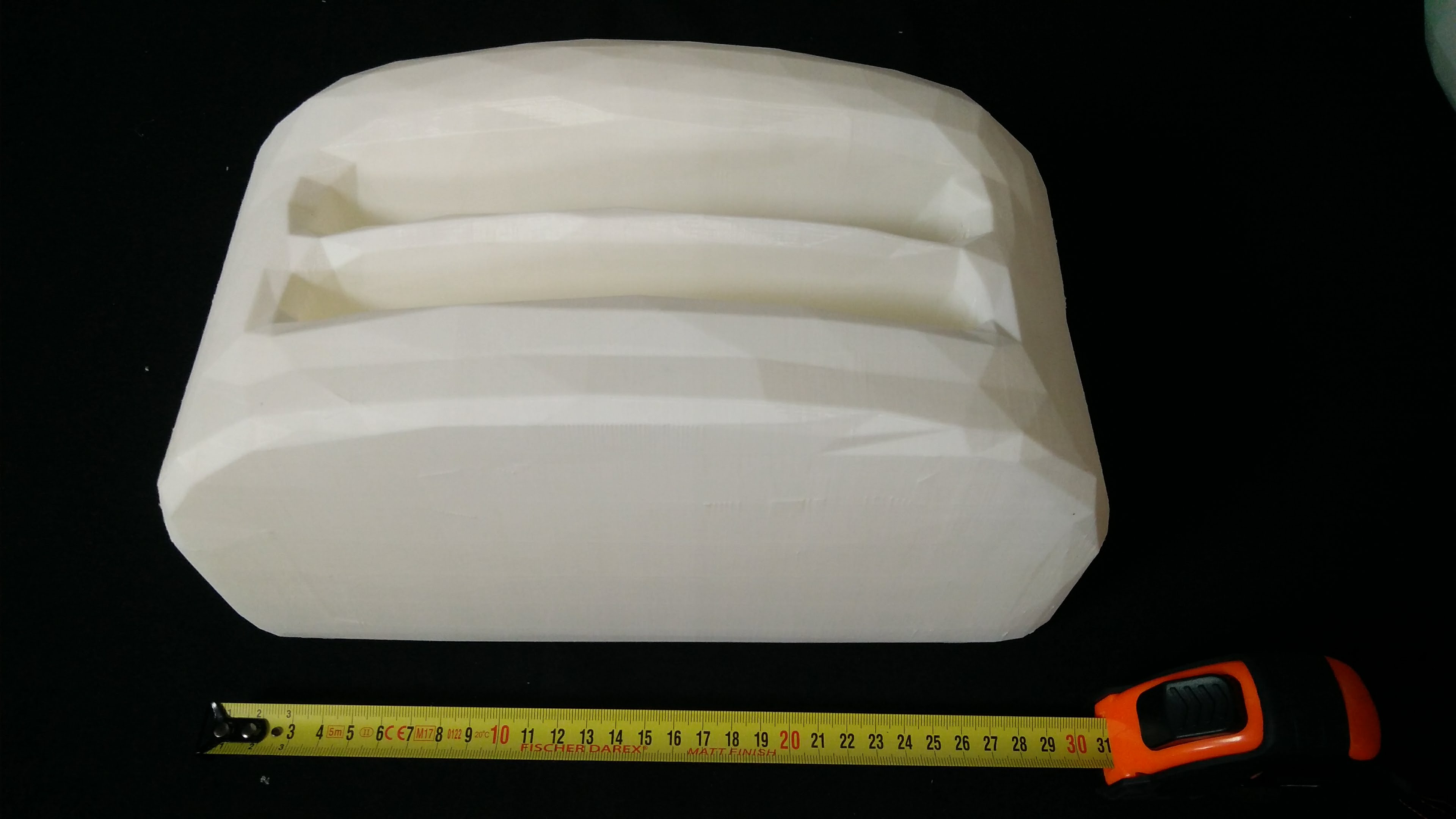}}
\captionsetup[subfloat]{width=3cm}
\subfloat[][Jar\_002]{\includegraphics[width=.32\columnwidth]{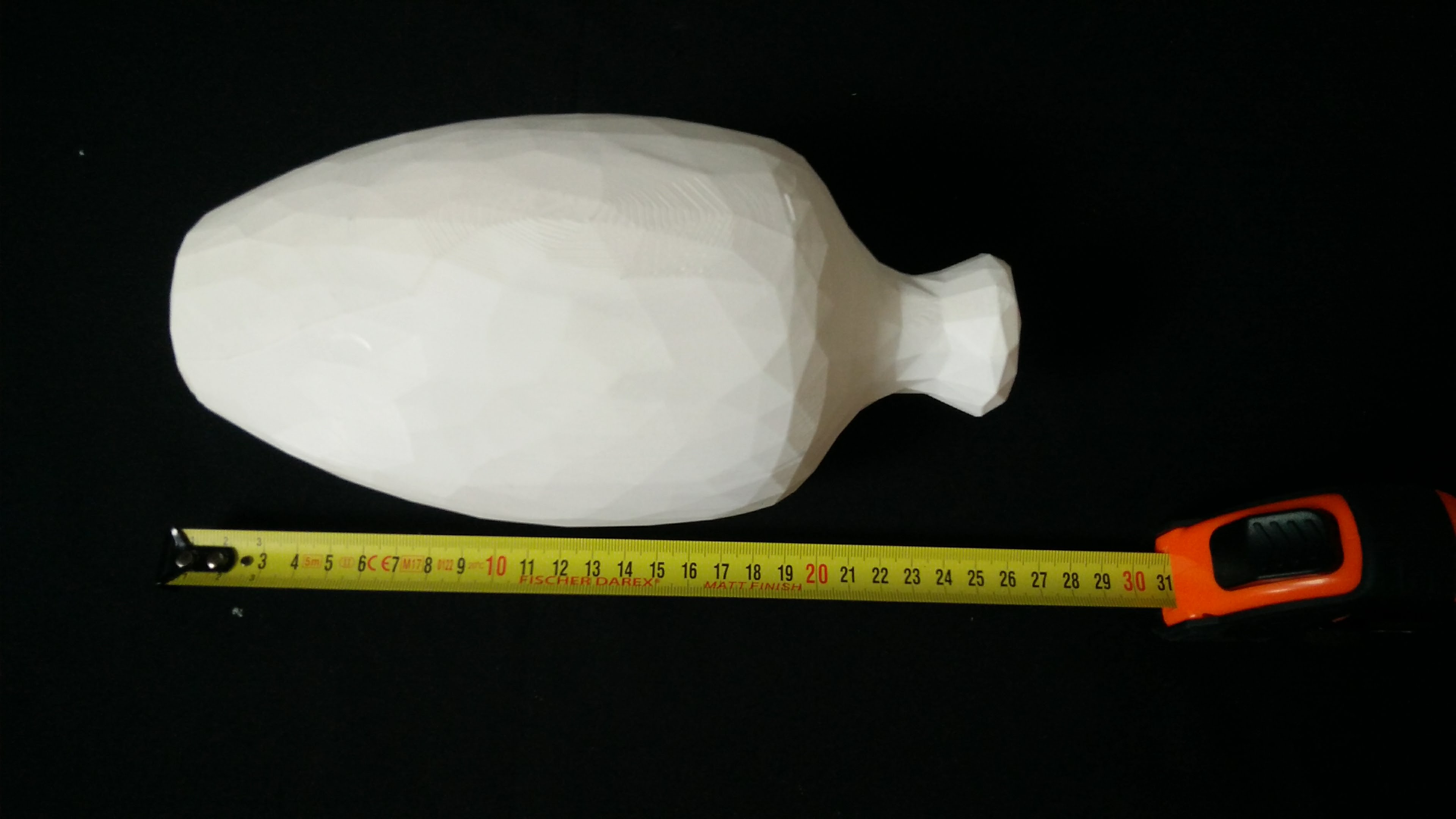}}
\captionsetup[subfloat]{width=3cm}
\subfloat[][Jar\_004]{\includegraphics[width=.32\columnwidth]{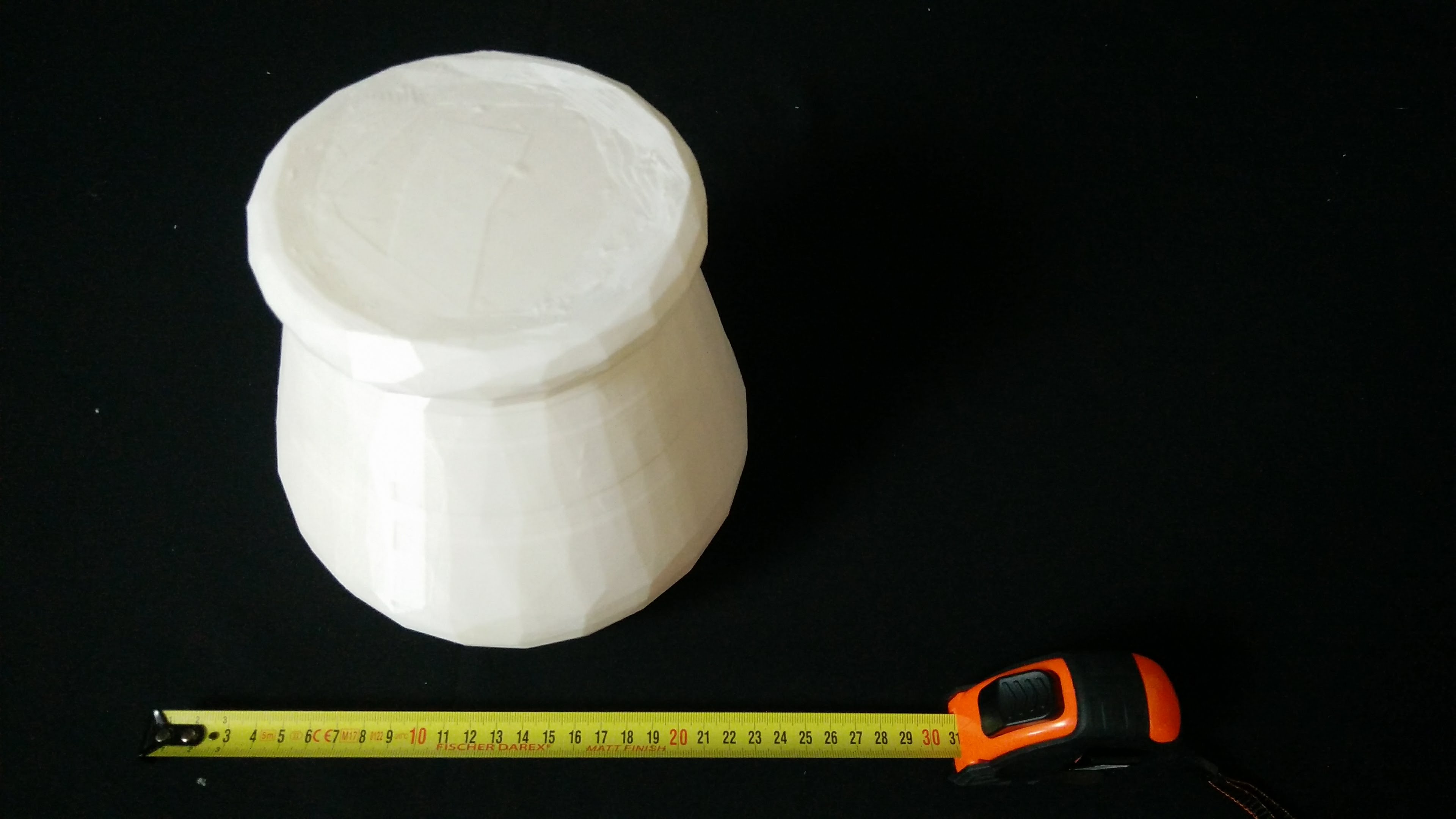}}
\caption{Object models used for the experiments with the Apollo robot system. Objects are: (a)Bottle 1, (b)Bottle 2, (c)Camera, (d)Lemon, (e)Bowl 1, (f)Bowl 2, (g)Toaster, (h)Jar 1 and (i)Jar 2. Subcaptions show the object name in the dataset.}
\label{fig:obj_real}
\end{figure}

Our experiments with the Apollo platform consider up to 9 different object models. These objects are printed using 3D printers and cover different weights, dimensions and shapes. The objects are: bottle 1, bottle 2, toaster, camera, lemon, bowl 1, bowl 2, jar 1 and jar 2. Figure \ref{fig:obj_real} shows the object models.

In the experiments with the Tombatossals robot we use only two different object models. Both objects were printed twice, using different percentages for the \textit{infill}. This provided identical object models but with different weights (light and heavy). Objects printed with different weights are bottle 1 and toaster.

\subsection{Grasp simulation}

All the grasps performed in real platforms are generated using OpenHand \cite{Leon2013}. Each grasp executed on real world is simulated and then evaluated using the quality metrics detailed in the previous section.

\section{Methodology}
\subsection{Experimental protocol}
To obtain the \textit{experimental score} of a candidate grasp, it is necessary to evaluate it on a real robotic platform. For the purpose of the experiments we will use only one arm and one hand on each robotic platform. To perform the experiments, we apply the following experimental protocol:

\subsubsection{Step 1: Initial setup. Move the arm/gripper to an initial pose.} The gripper is placed initially in a top-left position. This way the robot is able to plan and move the arm to a variety of poses over the table for grasping an object.

\subsubsection{Step 2: Detect and track the object pose.} To recognize and track the object we use the \textit{Depth-Based Bayesian Object Tracking Library} \cite{Wuthrich2013,Jan2016}. This library implements two different algorithms for object tracking: a particle filter and a Gaussian filter. In our experiment the particle filter is used. This library allows to automatically detect the desired object and obtain its pose w.r.t. the robot platform. 

\subsubsection{Step 3: Select a candidate grasp to be tested.} Among the different candidate grasps generated and evaluated with OpenHand, the human operator selects one to be tested. A candidate grasp has to be feasible to be selected. This means the robot or hand won't collide with the table when trying to acquire the grasp pose and the arm/planner is capable to move to that pose.
 
\subsubsection{Step 4: Move the arm/gripper to the grasp target pose.}
Once we have selected a feasible grasp, the planner moves the
arm/gripper to desired grasp target pose. 

\subparagraph{Exceptions:} There are some circumstances which may diverge the experimental protocol: the arm/gripper hits the object during the movement, the object pose is unstable prior to grasping, the object falls while the hand closes, the planner moves to a wrong pose or the object tracking fails/loses the object.  

\subsubsection{Step 5: Close the hand.} If the gripper achieves the desired grasp target pose, the robot starts closing its fingers until a minimum strain is detected for each finger joint on the gripper. Each joint has a strain threshold to ensure the gripper applies enough pressure over the object and not just touches it.

\subsubsection{Step 6: Move the gripper up: 15cm for small/medium objects, 25 cm for larger objects.} Once the minimum strain on each joint is achieved, the joints of the fingers are blocked and the arm starts moving the gripper between 15 to 25cm up in the air. If the grasp is stable the object will be lifted. For unstable grasps, the object will not be lifted or will slip during this lifting event.

\subsubsection{Step 7: \textbf{Hold the hand in the lift pose for three seconds.}} 
\begin{figure}[h]
\begin{center}
\includegraphics[width=.32\columnwidth]{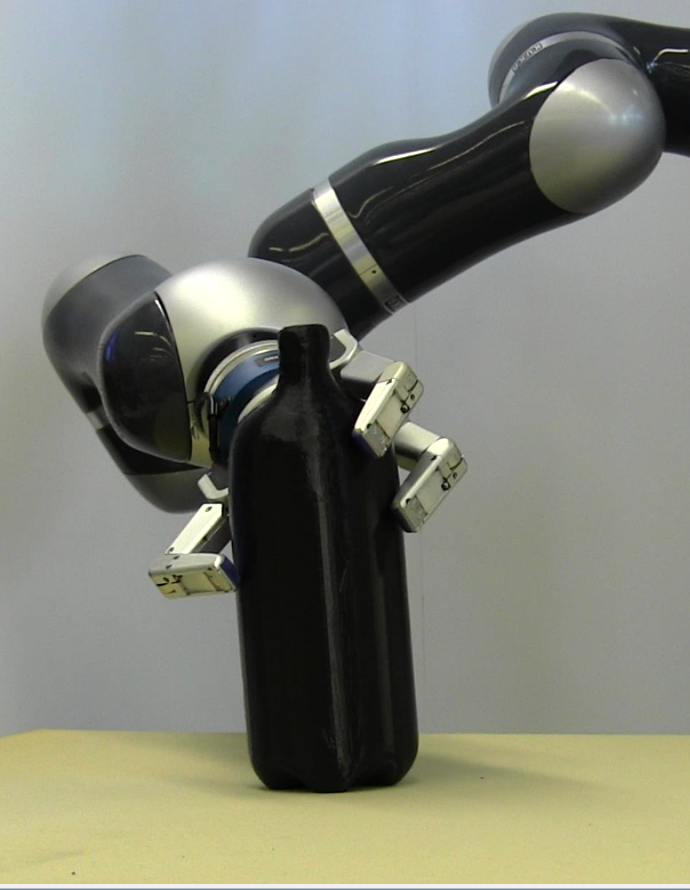}
\includegraphics[width=.32\columnwidth]{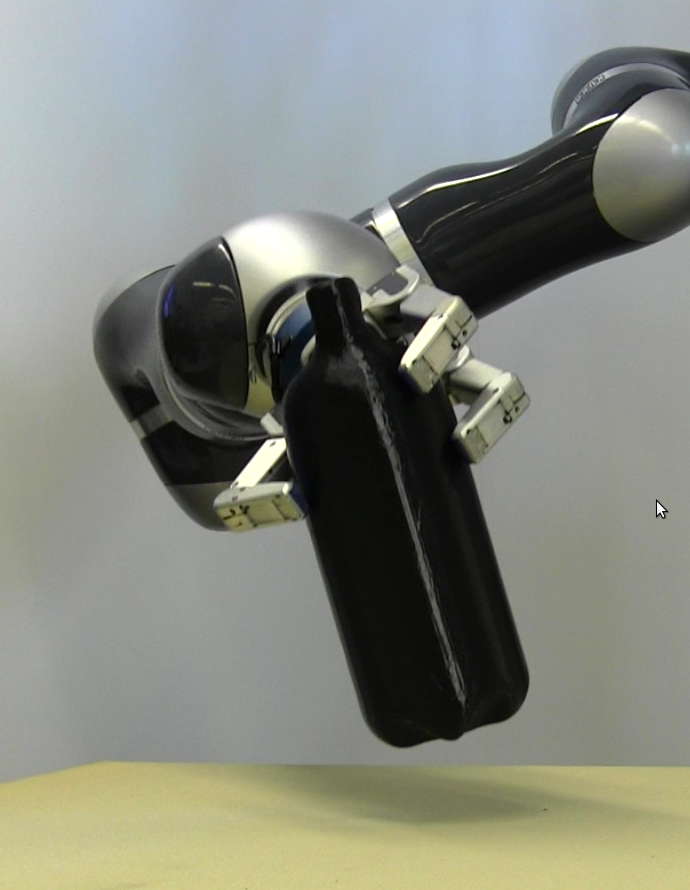}
\includegraphics[width=.32\columnwidth]{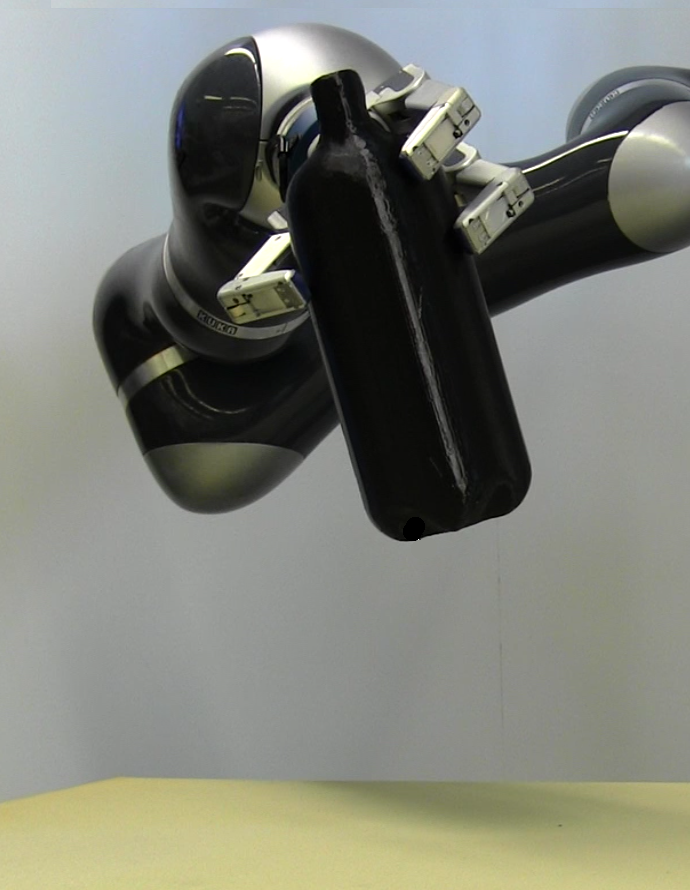}
\caption{Example of a successful real grasp execution }
\label{fig:real_suc}
\end{center}
\end{figure}

\begin{figure}[h]
\begin{center}
\includegraphics[width=.32\columnwidth]{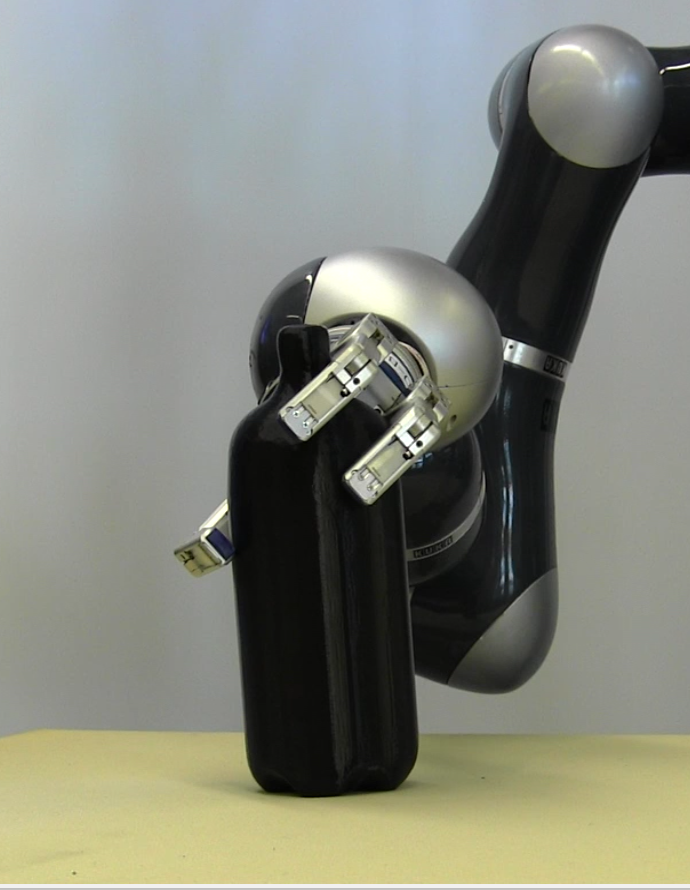}
\includegraphics[width=.32\columnwidth]{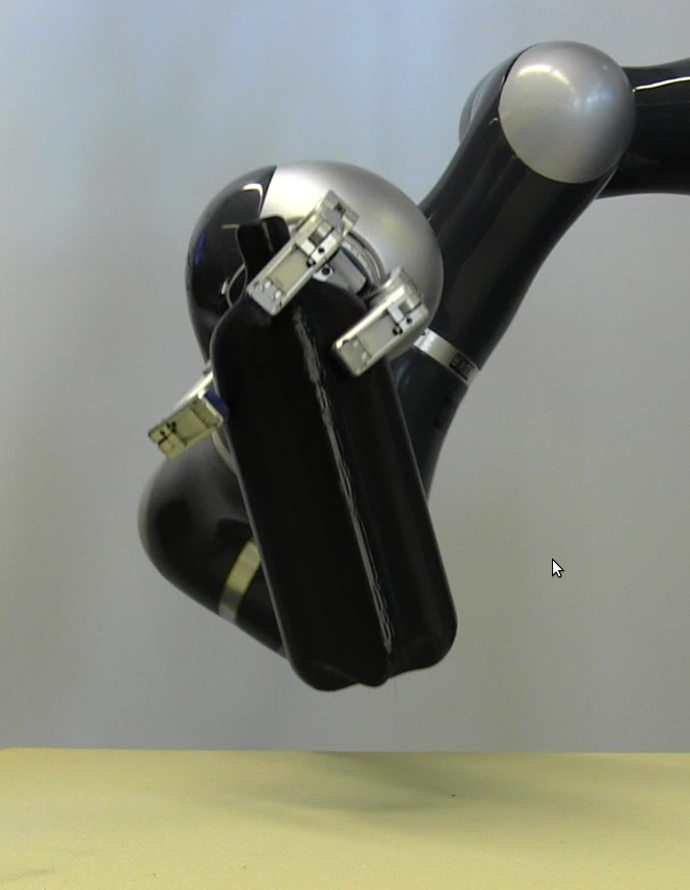}
\includegraphics[width=.32\columnwidth]{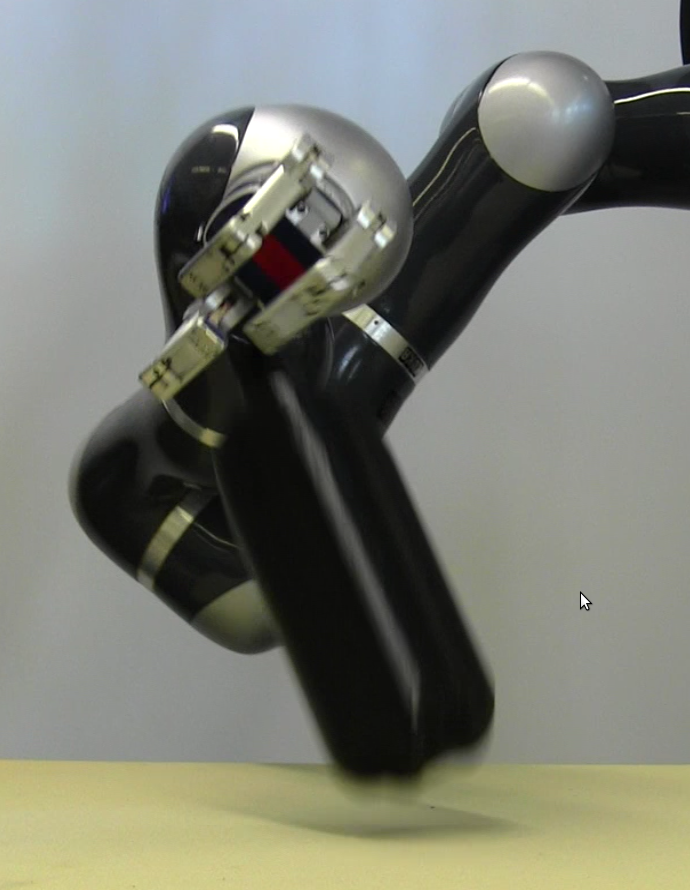}
\caption{Example of an unsuccessful real grasp execution }
\label{fig:real_un}
\end{center}
\end{figure}

Once the lift pose is achieved. The arm keeps this pose for 3 seconds, if the object remains in the gripper for this time, the grasp is considered \textit{Stable}. If this time has expired and the object is not in the gripper, the grasp is considered \textit{Unstable}. Figures \ref{fig:real_suc} and \ref{fig:real_un} illustrates two examples of successful and unsuccessful experimental grasps. Both experimental grasps correspond to the same candidate grasp.

\subsubsection{Step 8: Place the object on the table.} After holding the object, the gripper is placed again on the desired grasp pose. A margin of +2cm is applied in the \textit{Z axis} to avoid the object hitting directly the table surface.

\subsubsection{Step 9: Release the object.} After achieving the releasing pose, the fingers are opened and the object is finally released (in case the grasp was Stable).

\subsubsection{Step 10: Move the arm/gripper to the initial pose.} Finally the arm/gripper are moved to its initial pose (step 1).

\subsection{Grasp Classification}

Grasp executions on the real robot are initially scored as \textit{Stable} or \textit{Unstable}. After all the experiments are done, a post-processing is done for evaluating grasps using the 3-categories scale. If for a \textit{cluster of grasps} there are different experimental scores, all the grasps in the cluster are labeled as \textit{Fragile}, otherwise they are considered as \textit{homogeneous} and we keep their original score. 

A \textit{homogeneous} grasp could be either \textit{Robust} or \textit{Futile}. The term \textit{homogeneous} denotes a candidate grasp will always succeed or fail independent of the object's weight or the gravity orientation. 

\bigskip\statement{A grasp is considered \textbf{Fragile} if its success or failure is dependent  in the gravity orientation, object's weight or is highly dependent in the specific contact points.}

\bigskip\statement{A grasp is considered \textbf{Robust} if it always succeeds, independently of the gravity orientation, object properties or having accurate contact points.}

\bigskip\statement{A grasp is considered \textbf{Futile} if it always fails, independently of the gravity orientation, object properties or having accurate contact points.}

\begin{figure}[h]
\begin{center}
\includegraphics[width=\columnwidth]{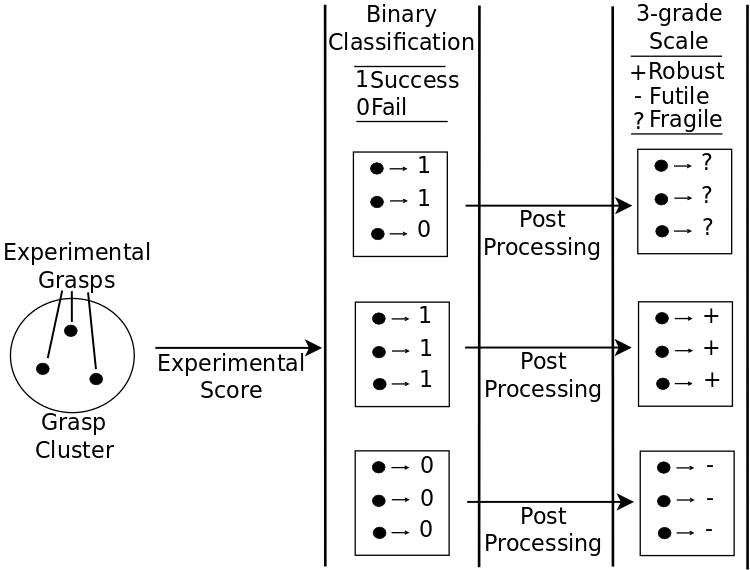}
\caption{Diagram for the labeling of experimental grasps using the binary classification method and the 3-grade scale. This diagram illustrates three examples of grasp clusters with different experimental scores.}
\label{fig:grasp_label}
\end{center}
\end{figure}

According to this, the classifiers and training applied consider a 3-Dimensional space where experimental grasps will be scored as \textit{Robust}, \textit{Futile} and \textit{Fragile}. Figure \ref{fig:grasp_label} shows the scoring of experimental grasps and whether \textit{Stable} or \textit{Unstable} grasps turn into \textit{Fragile}, \textit{Futile} or \textit{Robust} using the 2-grade scale system.

\section{Results}

First, Table \ref{tab:cs3_qm} shows the results of the training when considering only one metric as the input feature vector and the 3-categories score. Results are shown for the \textit{Tombatossals} grasp dataset the data is split randomly between train and test sets. 

\begin{table}[htbp]
\caption{Classification results for individual quality metrics. First column shows the 3-categories scale. Second column considers only \textit{Robust/Futile} grasps.} 
\label{tab:cs3_qm}
	\begin{center}
	\begin{tabular}{r|rr|rr}
	\hline
	 & \multicolumn{2}{r|}{3-categories scale} & \multicolumn{2}{r}{Robust vs. Fragile}  \\
	Metric & Train $\pm$ Std & Test & Train $\pm$ Std & Test \\ \hline
	$Q_{A1}$ & 0.67 $\pm$ 0.09 & 0.68 & 0.92 $\pm$ 0.06 & 0.85\\ 
	$Q_{B1}$ & 0.55 $\pm$ 0.04 & 0.56 & 0.72 $\pm$ 0.10 & 0.60\\ 
	$Q_{B2}$ & 0.54 $\pm$ 0.04 & 0.52 & 0.61 $\pm$ 0.07 & 0.63\\ 
	$Q_{B3}$ & 0.53 $\pm$ 0.01 & 0.48 & 0.57 $\pm$ 0.10 & 0.53\\ 
	$Q_{C2}$ & 0.55 $\pm$ 0.03 & 0.52 & 0.62 $\pm$ 0.07 & 0.53\\ 
	$Q_{D1}$ & 0.72 $\pm$ 0.04 & 0.73 & 0.90 $\pm$ 0.05 & 0.85\\ 
	$Q_{D2}$ & 0.51 $\pm$ 0.02 & 0.50 & 0.65 $\pm$ 0.10 & 0.60\\ \hline
	\end{tabular}
	\end{center}
\end{table}

Next, we analyze the performance of the classifiers using as input feature vector the full set of 7 quality metrics. We compare two methods to label grasps execution, the 3-category scale and a more simple binary classification: \textit{stable} or \textit{unstable} grasps. Table \ref{tab:cs3_comb} shows the performance comparing the 3-categories scale and the binary classification. 

\begin{table}[htbp]
\begin{center}
\caption{Comparison between the classification models for the 3-categories scale and the Binary Classification method. Combined datasets from Apollo and Tombatossals experiments.} 
\begin{tabular}{l|ll|ll}
\hline
 & \multicolumn{2}{c|}{Binary Classification} & \multicolumn{2}{c}{3-categories scale}\\
Classifier & Train$\pm$Std & Test & Train$\pm$Std & Test \\ \hline
K-Nearest Neighbors & 0.74$\pm$0.03 & 0.71 & 0.74$\pm$0.04 & 0.74 \\ 
Classification Trees & 0.72$\pm$0.05 & 0.69 & 0.76$\pm$0.04 & 0.76 \\ \hline
\end{tabular}
\label{tab:cs3_comb}
\end{center}
\end{table}

\section{Discussion}

First, if we analyze the results on predictive capability of metrics from Table \ref{tab:cs3_qm}, it is shown individual metrics have poor performance for predicting the stability of a grasp. The analysis on individual metrics showed that metrics tend to show a beter performance($90\%$) for evaluating only \textit{Robust/Fragile} grasps. However, they are still not sufficient individually for predicting the success of a grasp.

Results in real experiments showed that there is a clear difference between \textit{Robust}, \textit{Futile} and \textit{Fragile} grasps. First, our results show adding a new category for scoring the grasps: \textit{Fragile}, could improve the classification methods. Table \ref{tab:cs3_comb} showed this method is better than the binary scoring system

\section{Conclusions}

This study showed that individually, metrics are not good predictors, but they can be combined to improve their performance, up to an 75\% of accuracy. In our study, we assume we only have the information provided from the simulation for the classifiers. This is, we only have information regarding the contact points, quality metrics and object shape.

Under these assumptions, a 3-categories scale system for grasp executions was proposed. With this scoring method, grasps can be classified as \textit{Robust}, \textit{Futile} or \textit{Fragile}, depending if the execution of the candidate grasp always \textit{succeeds}, \textit{fails} or both.

This 3-categories scale method improved the classification and predictive capability of the models using quality metrics. Results showed this 3-categories scale reflects better the outcome of real grasp executions.

\subsection{Limitations}
This article presented an study on the performance of different quality metrics for predicting grasp success, but it had also some limitations. First, only two manipulators were used to perform the experiments and evaluate grasps. Extending the results on prediction models to other grippers should be done carefully, as results may vary. Second, we used a reduced number of objects to perform the experiments, an extended study with more objects should be done. 

Third, we considered different geometric characteristics of the objects, but there are other properties that should be taken in account: materials, elasticity, friction coefficient, etc. Fourth, the real grasps executions were restricted to an environment with a table holding the object prior the grasp. Repeating these experiments in other environments with different restrictions or without restrictions is advisable. Finally, the prediction models were generated using only two different types of classification methods. A wider study with this data can be done using other algorithms or methods, as it could provide better results.

\appendices



\ifCLASSOPTIONcaptionsoff
  \newpage
\fi

\bibliographystyle{IEEEtran}
\bibliography{references}


\end{document}